\documentclass[letterpaper, 10 pt, conference]{ieeeconf}  

\IEEEoverridecommandlockouts                              

\usepackage{graphics} 
\usepackage{epsfig} 
\usepackage{pbox}
\usepackage{subcaption}
\usepackage{hyperref}
\usepackage{footnote}

\usepackage{color}
\newcommand{\Lisa}[1]{#1}
\newcommand{\yang}[1]{#1}

\title{\LARGE \bf
Deep Learning for Tactile Understanding From Visual and Haptic Data 
}

\author{Yang Gao$^{1}$, Lisa Anne Hendricks$^{1}$, Katherine J. Kuchenbecker$^{2}$ and Trevor Darrell$^{1}$
\thanks{\yang{*This material is based upon work supported by DARPA, Berkeley Vision and Learning Center, as well as the U.S. National Science Foundation (NSF) under grants 1426787 and 1427425 as part of the National Robotics Initiative. Lisa Anne Hendricks is supported by the NDSEG.}} 
\thanks{$^{1}$ Department of Electrical Engineering and Computer Sciences, 
		University of California at Berkeley, California, CA 94701, USA
        {\tt\small \{yg, lisa\_anne, trevor\}@eecs.berkeley.edu}}%
\thanks{$^{2}$ Haptics Group, GRASP Laboratory, Department of Mechanical Engineering and Applied Mechanics,
        University of Pennsylvania, Philadelphia, Pennsylvania 19104, USA.
        {\tt\small kuchenbe@seas.upenn.edu}}%
}

\begin{document}

\maketitle
\thispagestyle{empty}
\pagestyle{empty}

\begin{abstract}

Robots which interact with the physical world will benefit from a fine-grained tactile understanding of objects and surfaces. Additionally, for certain tasks, robots may need to know the haptic properties of an object before touching it.  To enable better tactile understanding for robots, we propose a method of classifying surfaces with haptic adjectives (e.g., compressible or smooth) from both visual and physical interaction data.  Humans typically combine visual predictions and feedback from physical interactions to accurately predict haptic properties and interact with the world. Inspired by this cognitive pattern, we propose and explore a purely visual haptic prediction model.  Purely visual models enable a robot to ``feel'' without physical interaction.  Furthermore, we demonstrate that using both visual and physical interaction signals together yields more accurate haptic classification.  Our models take advantage of recent advances in deep neural networks by employing a unified approach to learning features for physical interaction and visual observations. 
\Lisa{Even though we employ little domain specific knowledge, our model still achieves better results than methods based on hand-designed features.}
\end{abstract}

\section{Introduction}



Tactile understanding is important for a wide variety of tasks.  For example, humans constantly adjust their movements based on haptic feedback during object manipulation \cite{johansson2009coding}.  Similarly, robot performance is likely to improve on a diverse set of tasks if robots can understand the haptic properties of surfaces and objects.  A robot might adjust its grip when manipulating a fragile object, avoid surfaces it perceives to be wet or slippery, or describe the tactile qualities of an unfamiliar object to a human.  In this work, we explore methods to classify haptic properties of surfaces. Using our proposed methods for haptic classification, we believe haptic information can be more effectively harnessed for a wide array of robot tasks.

Humans rely on multiple senses to make judgments about the world.  A well known example of this multisensory integration is the McGurk effect \cite{mcgurk1976hearing}, in which humans perceive different phonemes based on the interaction of visual and auditory cues.  Similarly, humans rely on both tactile and visual understanding to interact with their environment.  Humans use both haptic and vision to correctly identify objects \cite{woods2004visual}, and fMRI data demonstrates that haptic and visual signals are processed in a multi-sensory fashion during object recognition \cite{amedi2005functional}.  Motivated by the cross-modal processing inherent in the human brain, we build a model that processes both haptic and visual input and demonstrate that this combination achieves higher performance than using either the haptic or visual input alone.

\begin{figure}[!t]
  \centering
  \includegraphics[width=\columnwidth]{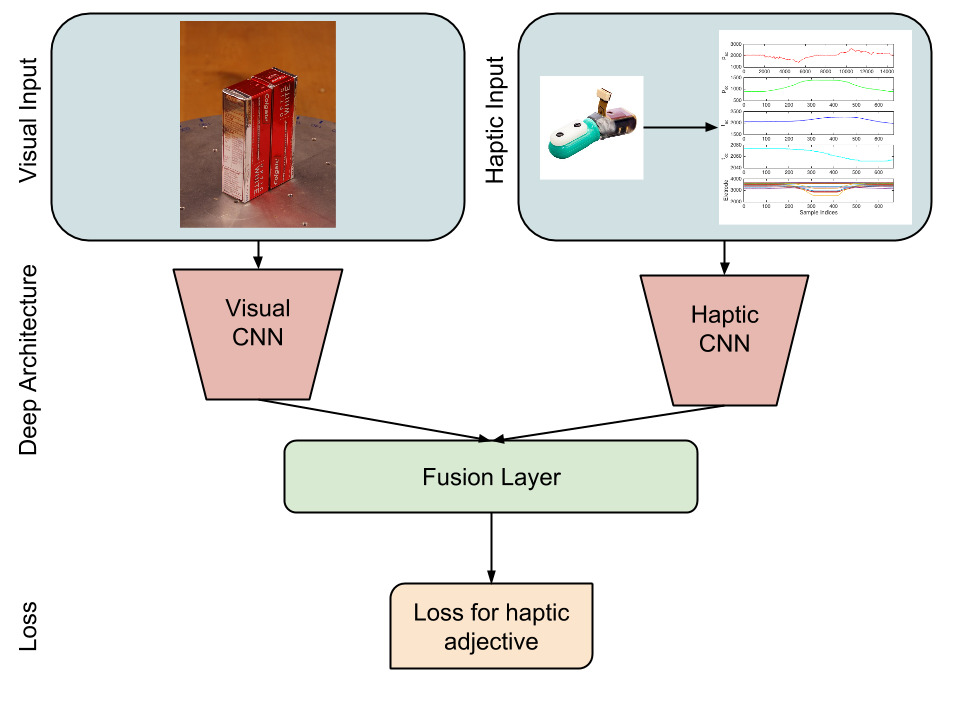}
  \caption[Caption for LOF]{We propose deep models as a unified approach to learning features for haptic classification of objects.  We use both visual and haptic data.  In comparison to training models on only haptic data or only visual data, we find that combining learned features across modalities, as pictured above, leads to superior performance. \footnotemark}
  \label{ConceptFigure}
\end{figure}

\footnotetext{To collect haptic data, we use the BioTac sensor \cite{wettels2008biomimetic}.  The image of the sensor in the above figure is courtesy of the product webpage \url{http://www.syntouchllc.com/Products/BioTac/}.}

In order to effectively learn from haptic and visual data, we train deep neural networks, which have led to dramatic improvements in disparate learning tasks such as object recognition \cite{krizhevsky2012imagenet} and automatic speech recognition \cite{hinton2012deep}.  As opposed to hand-designing features for particular modalities, neural networks provide a unified framework to learn features directly from data.  With our neural network model, we learn haptic features that outperform previous proposed features with little haptic domain knowledge.  Additionally, when training models on visual features, we transfer learned models on the related task of material classification to haptic classification.  Consequently, we can train a large visual model with less than 1,000 training instances. 

Our contributions are as follows. First, we demonstrate that neural networks serve as a unifying framework for signal classification, allowing us to learn rich features on both visual and haptic data with little domain knowledge.  We believe similar methods can be used to learn models for other signals that robots need to understand. Furthermore, we demonstrate that visual data from a different yet related task, material classification, easily transfers to haptic classification. Finally, we show that haptic and visual signals are complementary, and combining modalities further improves classification.

\section{Related Work}

\subsection{Previous Approaches to Tactile Understanding}

The tactile properties of consumer products have been explored to better understand the quality of items such as fabric \cite{pan2007quantification} and skin cream \cite{loden1992instrumental}.  Progress in tactile understanding has been driven by both the creation of better mechanical measurement systems (\cite{du2008fabric, wettels2008biomimetic}) as well as learning better algorithms for haptic understanding (\cite{du2008fabric, fishel2012bayesian}).  A variety of work  (\cite{fishel2012bayesian, mukaibo2005development}) concentrates on classifying specific textures (e.g., \textit{paper}).  Additionally, \cite{fishel2012bayesian} asserts that biomimicry is important for texture recognition.  In addition to using a BioTac sensor \cite{wettels2008biomimetic}, which is designed to replicate human tactile sensations, \cite{fishel2012bayesian} reproduces human movement when touching new textures.  Unlike these approaches, our system produces predictions of \textit{haptic adjectives}, enabling a qualitative description of previously unexplored textures and objects.

Prior work which focuses on haptic adjective classification includes \cite{griffith2012behavior}, \cite{Chu13-ICRA-Adjectives}, and \cite{Chu15-RAS-Adjectives}. \cite{griffith2012behavior} demonstrates that a rich and diverse haptic measurement system that measures temperature, compliance, roughness, and friction is key to accurately discerning between haptic adjectives such as \textit{sticky} and \textit{rough}.  Our work most closely resembles Chu et al.\ (\cite{Chu13-ICRA-Adjectives,Chu15-RAS-Adjectives}), which detail the collection of haptic classification datasets (PHAC-1 and PHAC-2) and concentrates on classifying objects with binary haptic adjectives.  \cite{Chu13-ICRA-Adjectives} and \cite{Chu15-RAS-Adjectives} rely on hand-designed features for haptic classification.  Two types of features are proposed: static and dynamic.  Static features consist of simple data statistics while dynamic features are learned from fitting haptic signals to HMMs.  

The haptic classification problem is closely related to the material classification problem. For example, classifying a surface as \textit{glass} implies the presence of some haptic properties, such as \textit{hard}, \textit{slippery}, \textit{smooth} and \textit{solid}. However, notable exceptions exist. For example, different plastic surfaces have vastly different roughness and hardness properties: a plastic bag is \textit{smooth} and \textit{soft} but a sawed plastic block is \textit{rough} and \textit{hard}. Consequently, haptic classification goes beyond simply identifying object materials.  \cite{bell2014material} details the collection of a large material classification dataset and then demonstrates that  deep models can be used for the material classification problem.  

\subsection{Neural Networks}

Neural networks have led to state-of-the-art results on many important problems in artificial intelligence (\cite{krizhevsky2012imagenet}, \cite{hinton2012deep}).  Neural networks are compositional models which are formed by stacking ``layers'' such that the output of one layer is the input of the next layer.  A layer consists of an affine transformation of layer inputs and learned weights followed by a nonlinearity.  Other operations such as max-pooling and normalization can also be included.  Neural networks are trained iteratively using backpropagation.  Backpropagation applies the chain rule in order to determine $\frac{dL}{dw_l}$ where $w$ are the weights of layer $l$ and $L$ is the loss function being minimized.  At each iteration, the gradients $\frac{dL}{dw_l}$ are used to update the weights in order to minimize loss.  

Convolutional neural networks (CNNs) \cite{lecun1998gradient} use a convolution as the linear operation within the layer, as opposed to fully connected layers (also called inner product layers) which use inner products as the linear operation within the layer.  CNNs have been proposed to decrease the total number of parameters of a model.  The reduction in the number of parameters has proven crucial for training large-scale object recognition systems \cite{krizhevsky2012imagenet} and has also been used for learning dynamics in 1-D time series, such as in speech recognition \cite{abdel2012applying}.  

For understanding 1-D time series, recurrent networks have also achieved great success on tasks such as language modeling \cite{mikolov2010recurrent} and handwriting recognition \cite{graves2008unconstrained}.  Recurrent networks include hidden units which are updated at each time step  based on the values of the hidden units at the previous time step and the current input.  Certain types of recurrent networks, such as LSTMs \cite{hochreiter1997long} and GRUs \cite{bahdanau2014neural}, include special gates which learn to remember and forget previous states.  Exemplary results on tasks such as machine translation \cite{bahdanau2014neural}, automatic speech recognition \cite{graves2014towards}, and parsing \cite{vinyals2014grammar} demonstrate that networks which include such gates are able to learn complex temporal dynamics, making them a natural fit for modeling time-series data.
 
In addition to performing better than other methods, neural networks are also advantageous because they require little domain knowledge.  This benefit is particularly appealing in the domain of haptics because high-dimensional haptic signals can be difficult for humans to understand.  Furthermore, weights learned from one dataset can be transferred to similar datasets through fine-tuning.  When fine-tuning, the model is initialized with weights learned on a previous task before training.  For example, weights from models trained on the large ImageNet dataset \cite{deng2009imagenet} can be transferred to related tasks such as semantic segmentation \cite{long2014fully}, scene classification \cite{zhou2014learning} and object detection \cite{girshick2014rich}. Pre-trained networks can also be used as off-the-shelf feature extractors by using the outputs (activations) of one of the layers of the network as a feature.  When a network has been pre-trained on a related task, using the network as a feature extractor leads to good results on tasks such as subcategory recognition and scene classification \cite{donahue2013decaf}.

Multimodal networks have been proposed in both unsupervised (\cite{ngiam2011multimodal}) and supervised (\cite{gupta2014learning, kahou2015emonets}) settings.  Both \cite{gupta2014learning} and \cite{kahou2015emonets} first train deep models on individual data modalities then use activations from these models to train a multimodal classifier.

\section{Deep Haptic Classification Models}

Deep learning provides a unified framework for learning models for classification.  We describe models that are able to achieve high performance on the haptic classification task using haptic data, visual data, and in a multimodal setting.

\subsection{Haptic Signal}
We explore both CNN and LSTM models for haptic classification.

\subsubsection{Haptic CNN Model}

\begin{figure}[thpb]
  \centering
    \includegraphics[width=\columnwidth]{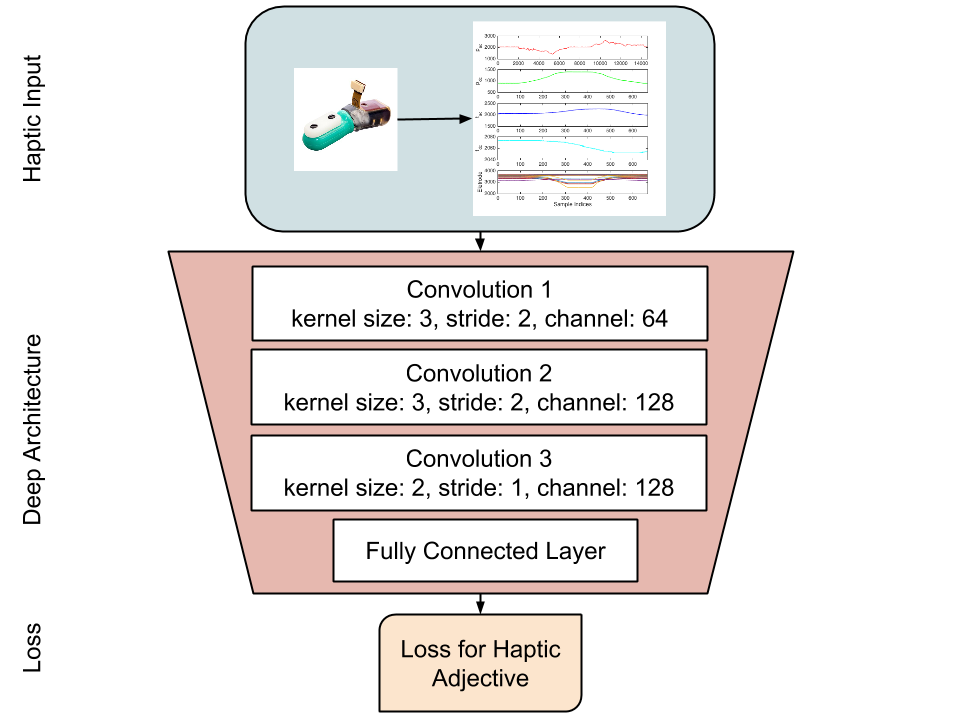} 
	\caption{Haptic CNN structure.
	}
	\label{fig_haptic_CNN}
\end{figure}

Our proposed CNN model performs temporal convolutions on an input haptic signal, similar to models which have previously shown good results on other tasks with 1-D signals, such as speech recognition \cite{abdel2012applying}.  CNNs decrease the total number of parameters in the  model, making it easier to train with limited training data.    

Fig. \ref{fig_haptic_CNN} shows the design of our CNN. Our haptic signal includes 32 haptic measurements (see~\ref{sec-preprocessing} for details).  To form the input of our network, we concatenate signals along the \textit{channel} axis as opposed to the \textit{time} axis.  Thus, concatenating two signals with dimension $T \times C$ where $T$ is the number of time steps and $C$ is the number of channels, results in a $T \times 2C$ signal.  After each convolution, we use a rectified linear unit (ReLU) nonlinearity.


To further reduce the number of parameters, we use ``grouping". Grouping for a certain layer requires that groups of channels do not interact with other groups.  We use groups of 32 in all our convolutional layers, which means interactions between different haptic signals are not learned until the fully connected layer. Generally, allowing cross-channel interactions earlier in the model is beneficial, and we anticipate better results without grouping when using a larger training dataset.

Our initial classification models are trained using logistic loss.  However we find that after learning the weights for the convolutional layers, fine-tuning with hinge-loss obtains similar or slightly better results for all models.  Unless otherwise mentioned, all reported results are from a network trained with a hinge loss.

\subsubsection{Haptic LSTM Model}
In addition to designing a CNN for haptic classification, we also explore LSTM models.  Because LSTM models have a recurrent structure, they are a natural fit for understanding haptic time-series signals.  Our LSTM structure consists of 10 recurrent units, and is followed by a fully connected layer with 10 outputs and a ReLU nonlinearity.  A final fully connected layer produces a binary prediction. We found that small changes in the LSTM structure (e.g. number of recurrent units) did not result in large changes in final classification scores.  Though stacking LSTMs generally leads to better results \cite{graves2014towards}, this led to performance degradation for haptic classification. 

\subsection{Visual CNN Model}

In addition to the haptic signal, we also use visual cues for haptic classification.  Since the haptic classification problem is closely related to material classification, we transfer weights from a CNN that is fine-tuned on the Materials in Context Database (MINC) \cite{bell2014material}.  MINC is a material recognition dataset, which consists of 23 classes (such as \textit{brick} and \textit{carpet}) and uses the  GoogleNet architecture \cite{szegedy2014going}. 
We refer to the fine-tuned MINC network as MINC-CNN.



\begin{figure}[thpb]
  \centering
  \includegraphics[width=\columnwidth]{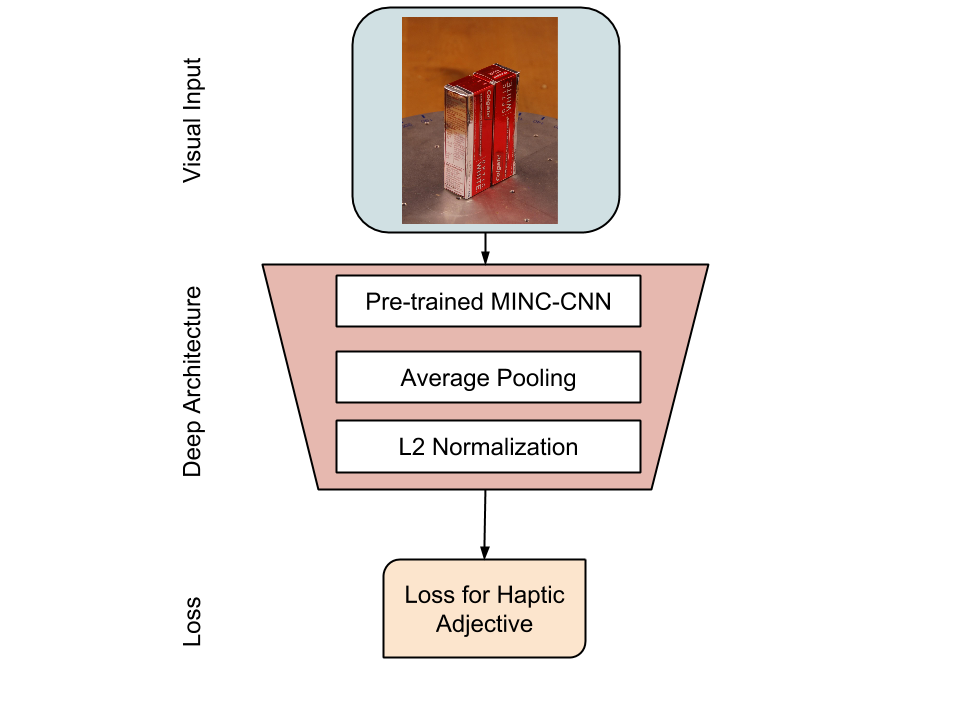}
  \caption{Visual CNN structure.}
  \label{fig_image_pipeline}
\end{figure}

To fine-tune a haptic classification model, we transfer weights for all layers below the ``inception (5a)"\footnote{The GoogleNet architecture is built from ``inception'' modules which consist of convolutions and filter concatenations.  Please see \cite{szegedy2014going} for more details.} layer of the MINC-CNN.  We find that placing an average-pooling layer and L2 normalization layer after the ``inception (5a)" and before a loss layer yields best results. Fig. \ref{fig_image_pipeline} summarizes the visual haptic adjective classification pipeline.

\subsection{Multimodal Learning}
 Figure~\ref{ConceptFigure} shows the structure of our multimodal model. The top left and top right denote the visual and haptic inputs respectively.  The learned weights from visual and haptic CNN models are transferred to the multimodal network.   Activations from the 3rd convolutional layer (conv3) of the haptic CNN and the activations after the L2-normalization of the inception (5a) layer of the visual CNN are concatenated.  The final multimodal classification model is trained with a hinge loss.  Note that when forming our multimodal network, we directly transfer weights from our previously trained haptic and visual networks, and only learn the classification layer.   

\subsection{Training method}
We use standard backpropagation to train our networks. We use SGD with a constant learning rate of $0.01$, momentum of $0.9$ and initialize model weights with ``xavier'' initialization \cite{glorot2010understanding}.  We train our network for 200 epochs with a batch size of 1000 (roughly $\frac{1}{5}$ of the entire training dataset).

\section{Experimental Setup}
\subsection{PHAC-2 Dataset}
\begin{figure*}[thpb]
  \centering
  \includegraphics[width=\textwidth]{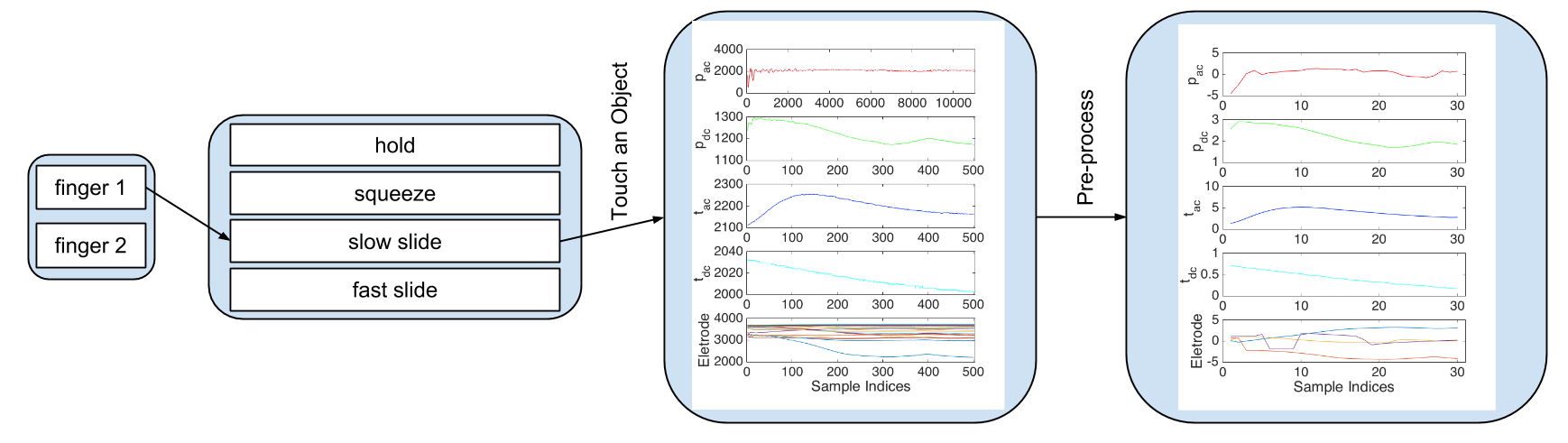}
  \caption{A sample BioTac signal of the first finger in the \textit{slow slide} exploratory procedure. Both raw signal and preprocessed signals are shown. The signals are $p_{AC}$, $p_{DC}$, $t_{AC}$, $t_{DC}$ and \textit{electrode impedance} respectively from top to bottom.}
  \label{fig_biotac_sample}
\end{figure*}

We demonstrate our approach on the Penn Haptic Adjective Corpus 2 (PHAC-2) dataset, which appears in \cite{Chu15-RAS-Adjectives}. 
The PHAC-2 dataset contains haptic signals and images of 53 household objects. Each object is explored by a pair of SynTouch biomimetic tactile sensors (BioTacs), which are mounted to the gripper of a Willow Garage Personal Robot 2 (PR2). Because the BioTac sensor mimics humans tactile capabilities, we believe its signals provide the rich haptic data necessary for fine-grained tactile understanding.  Each object is felt with the following four exploratory procedures (EPs): \textit{Squeeze, Hold, Slow Slide, Fast Slide}, which mimic how humans explore the tactile properties of objects. The BioTac sensor generates five types of signals: low-frequency fluid pressure ($P_{DC}$), high-frequency fluid vibrations ($P_{AC}$), core temperature ($T_{DC}$), core temperature change ($T_{AC}$), and 19 electrode impedance ($E_1 \dots E_{19}$) which are spatially distributed across the sensor (example signals in Fig.~\ref{fig_biotac_sample}). The $P_{AC}$ signal is sampled at 2200\,Hz, and the other signals are all sampled at 100\,Hz. Ten trials of each EP are performed per object. Although data on the joint positions and gripper velocity and acceleration are available, we concentrate on classifying the tactile signals.


\begin{figure}
    \centering
    \begin{subfigure}[b]{0.4\textwidth}
        \centering
         \includegraphics[width=0.8\textwidth]{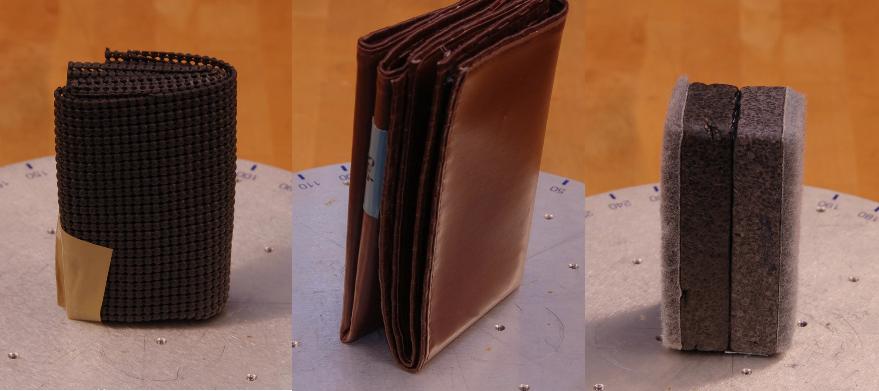}
        \caption{Examples of images automatically cropped by our algorithm described in~\ref{sec-visual-preprocessing}.}
        \label{example_objects}
    \end{subfigure}
    ~ 
    \begin{subfigure}[b]{0.4\textwidth}
        \centering
        \includegraphics[width=\textwidth]{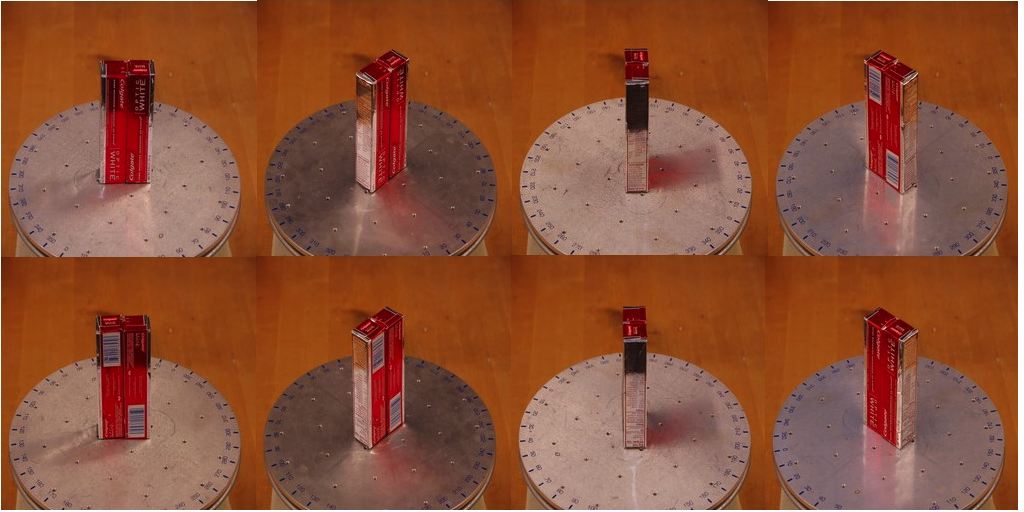}
        \caption{Each object is photographed from 8 different viewpoints.}
        \label{example_viewpoints}
    \end{subfigure}
    ~ 
    \caption{Visual data.}\label{visual_data}
\end{figure}

The dataset also contains high resolution ($3000\times 2000$) images of each object from eight different viewpoints (see Fig~\ref{visual_data}). The objects are placed in the center of an aluminum plate. Although lighting conditions are not explicitly controlled, the variations are insignificant. 

Each object is described with a set of 24 binary labels (Table~\ref{tb_adjs}), corresponding to the existence or absence of each of the 24 haptic adjectives (e.g., \textit{slippery} or \textit{fuzzy}).  For example, the toothpaste box is given the classification \textit{smooth}. The binary label for a given object was determined by a majority vote of approximately twelve human annotators~\cite{Chu15-RAS-Adjectives}. 

\begin{table}[h]
\caption{24 haptic adjectives}
\label{tb_adjs}
\begin{center}
\begin{tabular}{|llllll|}
\hline
absorbent & bumpy    & compressible & cool  & crinkly & fuzzy      \\
hairy     & hard     & metallic     & nice  & porous  & rough      \\
scratchy  & slippery & smooth       & soft  & solid   & springy    \\
squishy   & sticky   & textured     & thick & thin    & unpleasant \\
\hline
\end{tabular}
\end{center}
\end{table}

\subsection{Data Preprocessing}
\label{sec-preprocessing}
\subsubsection{Haptic Signal}

\textbf{Normalization}
For each signal $ s \in$ $\{P_{AC}, P_{DC},$ $T_{AC}, T_{DC},$ $E_1, $ $ \dots, E_{19} \}$, we calculate the mean $\bar{s}$ and the standard deviation $\sigma$, and normalize the signal by $s'=\frac{s-\bar{s}}{\sigma}$.

\textbf{Subsampling}
Because $P_{AC}$ is sampled at a higher rate than other signals, we first downsample it to 100\,Hz to match the sampling rate of other signals.  The durations of signals for different objects are almost identical for the EPs \textit{Hold}, \textit{Slow Slide}, and \textit{Fast Slide} 
but the signal length of \textit{Squeeze} varies considerably among objects. 
To resolve the disparity in signal lengths, we downsample signals for each EP to a fixed length of 150 for simplicity. Though our subsampling scheme possibly discards important information, reducing data dimensionality is crucial for training deep models with such a small dataset.  

\textbf{PCA on Electrode Impedances}
There are 19 electrode impedances spatially distributed across the core of BioTac. We found that 4 principal components capture 95\% of the variations of $E_1\dots E_{19}$. Consequently, we transform the 19 dimensional electrode impedance signals to the first four coefficients of the principal components. PCA is done independently on each EP across all objects.  Because we have four other signals ($P_{AC}, P_{DC},$ $T_{AC}, T_{DC}$), four electrode signals after PCA, and four EPs, we have a total of $(4+4)\times4 = 32$ haptic signals which we use as input to our models.

\textbf{Training Data Augmentation}
\label{tex_augmentation}
The total number of training instances is only 530 (53 objects with 10 trials each).  We use data augmentation to avoid substantial overfitting during model training.  To augment our data, we treat the two BioTac sensors on PR2 as two distinct instances.  Furthermore, we sample five different signals by sub-sampling each signal at a different starting point.  After data augmentation, we increase the number of total instances to 5,300.

\subsubsection{Visual Signal}
\label{sec-visual-preprocessing}
We follow standard image pre-processing steps and subtract the mean values from the RGB image and resize our image to the fixed input size of the MINC-CNN network ($224 \times 224$).  Instead of using the entire image, we extract a central crop which includes the object.  The central crop is obtained by first detecting the circular aluminum plate by its color, then estimating the plate's center and radius $R$. Our final crop is a rectangular region of size $2R\times R$ with the center of the crop above the center of the circular plate by $R$.  More generally, the crop could be obtained by a general bounding box detection method, or simply by focusing the camera on the surface of interest. Fig.\ref{example_objects} shows examples of images cropped with this method.

\subsection{Combining Data Instances}
\label{data-instances}
Combining separate instances from a single sample is a common method to boost classification results.  For example, in object recognition, it is common to augment data by using different image crops and mirroring images.  Better test classification can be achieved by averaging model outputs across separate instances for a single image.

Likewise, we explore combining the ten haptic explorations of each object and the eight visual viewpoints for each object.  However, instead of averaging model outputs at test time, we concatenate activations  (conv3 for haptic signals and inception (5a) for visual signals) for different instances and retrain the network loss layer.




\subsection{Train/Test Splits}
Following \cite{Chu15-RAS-Adjectives, Chu13-ICRA-Adjectives}, we partition the 53 objects into 24 $90/10$ train/test splits, one for each adjective. Creating 24 separate train/test splits is necessary because there exists no one split such that each adjective is present in both the train and test split.  Note that although a single object has several corresponding measurements, we do not allow the same object to appear in both the train and test split.

\subsection{Performance Metric}
To measure the performance of the approaches proposed, we adopt the widely used Area Under Curve (AUC) metric. AUC measures the area under the Receiver Operating Characteristic curve, which takes both the true positive rate and the false positive rate into consideration. The previous methods report F1 score \cite{Chu15-RAS-Adjectives, Chu13-ICRA-Adjectives}, which is biased towards true positives \cite{powers2011evaluation}. To provide a fair evaluation, we reproduce the baseline methods and report AUC scores. 

\section{Evaluation}
 \label{sec_evaluation}

In Table \ref{tb_auc} we compare the AUC scores from our proposed methods.  We determine AUC scores for each adjective and report the average over AUC scores for all adjectives.  We report models trained on the haptic, visual, and multimodal data sources as well as models trained by combining features from different trials on a single object (see section~\ref{data-instances}).  To differentiate our models, we use the following notation: the prefix for each model denotes the data source on which it is trained, and the suffix (if present) denotes how many instances are combined for classification.  For example, ``Haptic-ConvNet-10'' is a CNN model trained on haptic data and concatenating the conv3 activations from 10 haptic trials before classification.  

Models A-F are SVMs trained with the shallow features from \cite{Chu15-RAS-Adjectives, Chu13-ICRA-Adjectives} \footnote{Code for our reproduction of this result and further evaluation details, including comparison of reproduced F1 numbers, are available at \url{http://www.cs.berkeley.edu/~yg/icra2016/}}.  Models G-J and K-L are our deep haptic and deep visual models.  Models M-N are our multimodal models, and model O is the result for a model that randomly guesses.

When comparing the features from previous work, we find that the static features (methods A, B) achieve a slightly higher AUC than dynamic features (methods C, D).  Combining static and dynamic features (method F) increases the AUC by $0.9$.  

Though the hand-designed features perform well, our best deep CNN trained on haptic data (model I) improves the AUC score by $5.0$.  In model H, we augment our data by treating trials from the two BioTac as separate training instances and by subsampling each signal starting a different time step and using each sample as a separate training instance.  This leads to a tenfold increase in the size of the training set (2 signals from each BioTac sensor and 5 subsampled signals).  However, even though there is more available training data, we achieve better results by concatenating features from both BioTac recordings and each subsampled signal (model I).  One explanation is that examples from each BioTac sensor and different subsampling schemes are extremely similar, causing the model to overfit. However, combining haptic signals for all trials for a given object (model J) results in slightly lower AUC compared to model I.  This could be because haptic signals from various trials have informative differences that enable the model to generalize better to signals on new objects.  We see the same general trend when using the ``shallow'' static and dynamic baseline features.  We also classify haptic signals by using our LSTM model (model G).  Though LSTMs generally are quite good at learning complex temporal dynamics, in the haptics domain they perform considerably worse than the haptic CNN model. \yang{Variations of LSTM, such as changing the inputs to the frequency domain, could potentially improve the results.  We leave further investigations to future work. } 

Our visual models also perform well, though performance is $6.0$ points below our best haptic model.  In contrast to our haptic models, we find that combining features across multiple viewpoints is essential for good haptic classification and improves AUC from $71.5$ to $77.2$.  For robotic applications, this implies that robots need to view objects from varied viewpoints in order to gain an understanding of haptic properties.

We find that our multimodal models perform the best by well over $2.7$ points.  When concatenating activations from our best haptic model (I) and best visual model (L), we achieve an AUC of $84.7$.  However, by concatenating activations from model J and model L, we achieve an AUC of $85.9$, even though the corresponding haptic only model performs slightly worse.



\begin{savenotes}
\begin{table}[h]
\caption{Comparison of Haptic Classification Methods. See section~\ref{sec_evaluation} for details. }
\label{tb_auc}
\begin{center}
\begin{tabular}{|c||l|c|}
\hline
 & Methods                   & AUC (\%)  \\
\hline
A      & Haptic-static-1 trial     & \textbf{77.3}  \\
B      & Haptic-static-10 trials   & 70.8 \\
C      & Haptic-dynamic-1 trial    & \textbf{75.5} \\
D      & Haptic-dynamic-10 trials  & 74.3 \\
E      & Haptic (Combine model A + C)         & 77.5 \\
F      & Haptic (Combine model A + D)         & \textbf{78.2} \\
\hline
G      & Haptic-LSTM               & 72.1  \\
H      & Haptic-ConvNet & 81.6 \\
I      & Haptic-ConvNet-1 trial    & \textbf{83.2} \\
J      & Haptic-ConvNet-10 trials  & 82.1 \\
K      & Image-GoogleNet-1 view    & 71.5 \\
L      & Image-GoogleNet-8 views   & \textbf{77.2} \\
M      & Multimodal (Combine model I + L)             & 84.7 \\
N      & Multimodal (Combine model J + L)         & \textbf{85.9} \\
\hline
O      & Random guess              & 50.0  \\
\hline
\end{tabular}
\end{center}
\end{table}
\end{savenotes}

We release our code in order to make our results easily replicable \footnote{Code and data are available at \url{http://www.cs.berkeley.edu/~yg/icra2016/}.}.  Additionally, by releasing our models, we hope our haptic classifiers can be integrated into current robotic pipelines that could benefit from better tactile understanding of real objects and surfaces.

\section{Discussion}

In order to gain a more intuitive understanding of our model, we present analysis to illustrate what our models are capable of learning.

\subsection{Adjective Prediction Using Different Models}
\label{sec_ana_samples}
In order to compare the types of predictions produced by our models, we randomly select three objects out of the 53 objects in the data set and examine haptic predictions for each object.  


Table~\ref{tb_analysis_samples} details the predicted adjective for three objects: \textit{shelf liner}, \textit{placemat}, and \textit{furry eraser} (see Fig~\ref{example_objects}).  We observe that the haptic classifier tends to have high recall, predicting many adjectives for each class.  In contrast,  the visual classifier is more conservative and appears to have higher precision.  For the three examples shown, the visual model predicts no false positives.



The prediction of the multimodal classifier is not a simple union of the haptic and visual classifiers, but rather combines the two sources of information sensibly. For example, the prediction of shelf liner contains the correct \textit{textured} label which does not occur in either the haptic or visual results. Consequently, the multimodal model has higher precision and recall than the haptic model and higher recall than the visual model.  For some objects, such as \textit{furry eraser}, the multimodal classifier performs worse than the haptic classifier.  However, the multimodal model generally performs better than either the haptic only or visual only model leading to higher overall AUC.

\begin{table*}[ht]
\caption{Comparison of Haptic Classification Methods. The letter in parentheses refers to the configuration in Table \ref{tb_auc}.}
\label{tb_analysis_samples}
\vspace{-.1in}
\begin{center}
\begin{tabular}{l|l|l|l}
\hline
                                  & Shelf liner                                 & Placemat                  & Furry eraser                              \\
\hline                                  
Ground Truth          & bumpy compressible squishy textured         & smooth                    & fuzzy squishy textured                    \\
Haptic Outputs (I)    & absorbent compressible soft springy squishy & compressible textured     & fuzzy hairy soft squishy textured         \\
Visual Outputs  (L)   & squishy                                     & smooth                    & (empty)                                   \\
Multimodal Outputs (M) & compressible scratchy squishy textured      & compressible smooth thick & compressible rough squishy textured thick\\
\hline
\end{tabular}
\vspace{-.2in}
\end{center}
\end{table*}

\subsection{Haptic and Visual Data are Complementary}
\begin{figure*}[thpb]
  \centering
  \includegraphics[width=\textwidth,clip,trim=0 .24in 0 0]{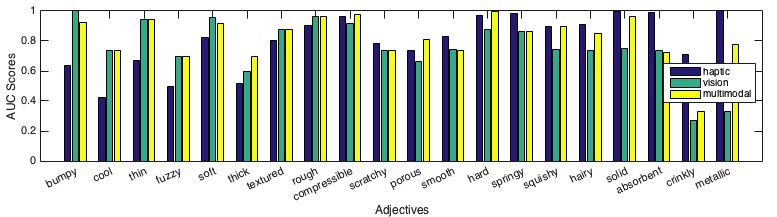}
  \vspace{-.2in}
  \caption{Comparison of AUC scores for our best visual and best haptic classifier for each adjective.  AUC scores are averaged over three train/test splits.}
  \vspace{-.15in}

  \label{fig_hap_vis_adj}
\end{figure*}

Although it seems most natural to classify haptic adjectives purely by touch, we demonstrate that visual and haptic signals are complementary.  Fig. \ref{fig_hap_vis_adj} compares the accuracy of our visual model to our haptic model.  Adjectives which describe object size such as \textit{thin} and \textit{thick} are better classified by visual data whereas adjectives such as \textit{squishy} and \textit{absorbent} are better classified by the haptic signal.  Surprisingly, even though the BioTac is equipped with a temperature sensor, our visual model is better at determining whether items are \textit{cool}.  Though the adjective \textit{metallic} might seem like something that would be understood using visual cues, we find that our haptic model is better at classifying objects as \textit{metallic} or not.  This could possibly be because material properties such as temperature and roughness are important in classifying \textit{metallic} objects, but not all metal surfaces have comparable metallic lusters.  For example, objects such as the toothpaste box in Fig \ref{example_viewpoints} appear shiny but are not actually \textit{metallic}.

\subsection{Which Haptic Signals Matter?}
Although designing a deep architecture that reliably classifies haptic signals requires little haptic domain knowledge, observing activations within the net can provide insight into what kinds of signals are important for classification.  In our design of the haptic CNN, we use a ``grouping'' strategy, such that each signal, such as $p_{AC}$ or $p_{DC}$, is processed independently before the fully connected layer. By looking at the activations of the final convolutional layer (conv3), we can observe which channels result in higher activations, and could thus be more important in haptic classification.

\begin{figure*}[thpb]
  \centering
  
    \begin{subfigure}[t]{0.24\textwidth}
        \centering
        \includegraphics[width=\textwidth]{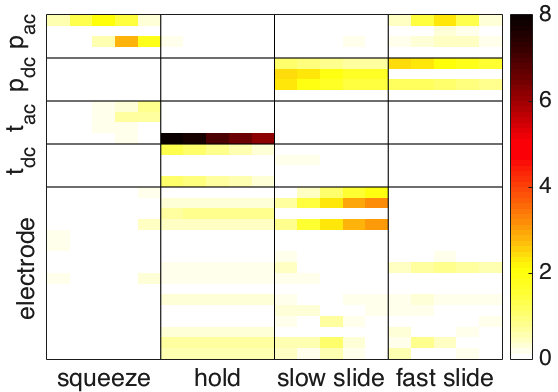}
        \caption{Adjective: metallic; Object: aluminum channel, Correct classification}
    \end{subfigure}%
    ~ 
    \begin{subfigure}[t]{0.24\textwidth}
        \centering
        \includegraphics[width=\textwidth]{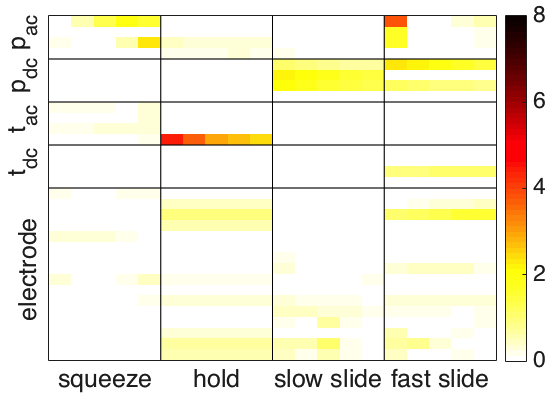}
        \caption{Same as (a) but activations from another trial. }
    \end{subfigure}%
    ~
    \begin{subfigure}[t]{0.24\textwidth}
        \centering
        \includegraphics[width=\textwidth]{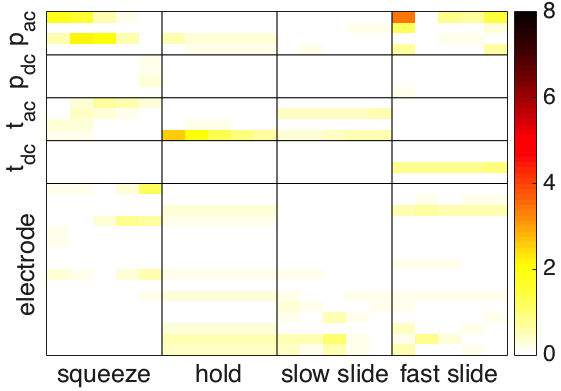}
        \caption{Adjective: metallic; Object: aluminum channel, Wrong classification}
    \end{subfigure}%
    ~
    \begin{subfigure}[t]{0.24\textwidth}
        \centering
        \includegraphics[width=\textwidth]{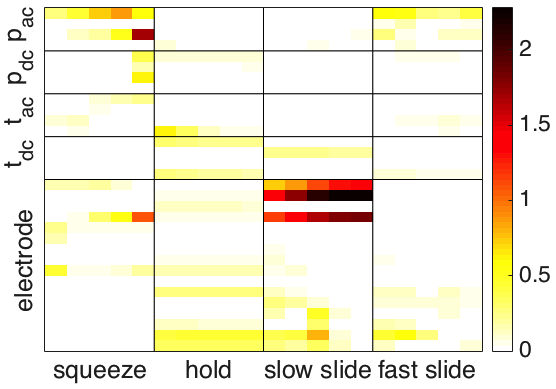}
        \caption{Adjective: compressible, Object: dish cloth, Correct classification}
    \end{subfigure}%
    \caption{Activations of  conv3 in our haptic classification pipeline. We show the activations for two objects with the \textit{metallic} and \textit{compressible} property respectively. The haptic classifier correctly predicts the \textit{metallic} adjective in (a) and (b), but not (c).  The classifier correctly predicts \textit{compressible} in (d).  Darker colors represent higher activations.} 
    

    
  \label{fig_activations}
\end{figure*}

In Fig. \ref{fig_activations}, we plot conv3 activations for 2 adjective classifiers (\textit{metallic} and \textit{compressible}). The activations are obtained on the held-out positive testing instances. We show two \textit{metallic} test instances which are classified correctly, and one \textit{metallic} test instance which is classified incorrectly.  Additionally, we show one test instance for \textit{compressible} which is classified correctly.  We observe that most of activations are zero, as many authors have observed \cite{agrawal2014analyzing} when analyzing networks trained on images. 

For the adjective \textit{metallic}, it appears activations for \textit{Core Temperature Change} ($t_{AC}$) signal are important for classification, which makes intuitive sense.  Because \textit{metallic} objects are presumably made of thermally conductive material, temperature is likely a valuable signal for haptic classification.  Furthermore, the activations of different trials look similar, as shown in Fig.~\ref{fig_activations} (a) and (b). This indicates the learnt haptic CNN is robust to signal variations across trials. Furthermore when the activations corresponding to $t_{AC}$ are not strongly activated, as in Fig.~\ref{fig_activations} (c), the signal is incorrectly labeled as not \textit{metallic}.  

For the adjective \textit{compressible} we note that electrode activations appear to be important.  This finding implies that different signal channels are important for classification of different adjectives, suggesting that most of the signals recorded by the BioTac sensor are important for haptic classification.  In future work, training models with data collected by less advanced sensors could verify the importance of the high dimensional BioTac sensor for fine-grained tactile understanding.


\section{Conclusion}
We successfully design deep architectures for haptic classification using both haptic and visual signals, reinforcing that deep architectures can be used as a paradigm to learn features for a variety of signals.  Furthermore, we demonstrate that haptic and visual signals are complementary, and combining data from both modalities improves performance.  We believe that integrating our models into robotic pipelines can provide valuable haptic information which will help boost performance for myriad robot tasks.

\Lisa{
In the future, we believe that our model can be improved by a larger, more diverse dataset.  Recently, \cite{burka2015toward} discussed efforts to collect a substantially larger dataset of paired visual and haptic data. A larger dataset could possibly improve our model in a variety of ways: it would allow to train a larger network, fine-tune through the entire multimodal network, and less aggressively down sample our haptic signal.
}







\section*{ACKNOWLEDGMENT}
\yang{We would like to thank Jeff Donahue for advice and guidance during the initial stages of the experiments, as well as for useful discussions on deep models.}




\bibliographystyle{IEEEtranS} 
\bibliography{root}

\end{document}